\documentclass[a4paper,twoside]{article}

\usepackage[breaklinks=true,bookmarks=false]{hyperref}
\usepackage{graphicx}
\graphicspath{ {./images/} }

\usepackage{epsfig}
\usepackage{subcaption}
\usepackage{calc}
\usepackage{amssymb}
\usepackage{amstext}
\usepackage{amsmath}
\usepackage{amsthm}
\usepackage{multicol}
\usepackage{pslatex}
\usepackage{apalike}
\usepackage{SCITEPRESS}     

\begin{document}

\title{Automated Damage Inspection of Power Transmission Towers from UAV Images}


\author{\authorname{Aleixo Cambeiro Barreiro\sup{1}\orcidAuthor{0000-0002-1019-4158}, Clemens Seibold\sup{1}\orcidAuthor{0000-0002-9318-5934}, Anna Hilsmann\sup{1}\orcidAuthor{0000-0002-2086-0951} and Peter Eisert\sup{1,2}\orcidAuthor{0000-0001-8378-4805}}
\affiliation{\sup{1}Fraunhofer HHI, Berlin, Germany}
\affiliation{\sup{2}Humboldt University of Berlin, Berlin, Germany}
\email{\{aleixo.cambeiro, clemens.seibold, anna.hilsmann, peter.eisert\}@hhi.fraunhofer.de}
}


\abstract{
    Infrastructure inspection is a very costly task, requiring technicians to access remote or hard-to-reach places. This is the case for power transmission towers, which are sparsely located and require trained workers to climb them to search for damages. Recently, the use of drones or helicopters for remote recording is increasing in the industry, sparing the technicians this perilous task. This, however, leaves the problem of analyzing big amounts of images, which has great potential for automation. This is a challenging task for several reasons. First, the lack of freely available training data and the difficulty to collect it complicate this problem. Additionally, the boundaries of what constitutes a damage are fuzzy, introducing a degree of subjectivity in the labelling of the data. The unbalanced class distribution in the images also plays a role in increasing the difficulty of the task. This paper tackles the problem of structural damage detection in transmission towers, addressing these issues. Our main contributions are the development of a system for damage detection on remotely acquired drone images, applying techniques to overcome the issue of data scarcity and ambiguity, as well as the evaluation of the viability of such an approach to solve this particular problem.
}

\onecolumn \maketitle \normalsize \setcounter{footnote}{0} \vfill

\section{\uppercase{Introduction}}
\label{sec:introduction}

Damage inspection is a core task in the maintenance process of big infrastructures. It is vital to detect structural issues as soon as possible in order to prevent further damages or even a complete collapse. Especially for critical infrastructures, damages need to be detected and evaluated before they cause any harm. However, this inspection can be a daunting task, since such structures are often located very sparsely, have a very large size and are comprised of numerous individual components. Additionally there is usually a large intra-class variability. All of these issues are present in inspections of power transmission towers, which need to be performed on a regular basis. The traditional way to do this is to have specialized technicians travel to the location of each of the towers in the energy network, deploy the necessary security equipment and inspect from every side, which can be very challenging depending on the environment. Helicopters are also sometimes additionally required to fly along the transmission lines. However, with the recent increase in availability of drones to the general public, the industry is moving towards remote acquisition of images of these structures. Feasibility of such approaches and working examples are documented in~\cite{morgenthal2014quality,sony2019literature}. This would allow a visual analysis of the components without the need of experts physically accessing them. Although it is already a big improvement with respect to physical inspection, this approach still has its limitations. One of them is the fact that all this information must be manually inspected, which is in itself a very costly task to perform.

Automation of the visual inspection addresses this issue. Vision-based detection techniques have improved vastly in recent years following the revolution in the field of machine learning brought by artificial neural networks. These methods rely on large datasets of labeled images to be used as training data, which results in another challenge: due to power transmission towers being critical infrastructure there are, to the best of our knowledge, no freely available datasets of labeled images for this task.  Therefore, images had to be collected and manually labeled for our experiment. The difficulty and effort associated with image collection and their labelling mean that only a very limited amount of data is available. This problem is aggravated by the fact that images with damages are rare, since these damages are repaired as soon as possible, making it impossible to gather a large dataset. Additionally, the exact definition of a damage and its extent are in some cases unclear, especially in areas further away from the camera and/or with low visibility, introducing a degree of subjectivity to the labeling process. Hence, one of the focal points of this paper is dealing with the scarcity of data and abundance of unclear examples.

Due to the aforementioned lack of training data, the amount of literature related to this problem is limited. Some examples of work that tackle similar issues are~\cite{gopalakrishnan2018crack,varghese2017power,shihavuddin2019wind}. Gopalakrishnan et al. also use deep-learning methods to detect damages in structures in images taken from unmanned aerial vehicles (UAV), but they only perform classification of the images and not localization of the actual damages~\cite{gopalakrishnan2018crack}. Varghese et al. as well apply deep-learning techniques to drone images of power transmission lines and towers, but the focus is mainly placed on detecting the structures: the only type of damage detected is tilting of the towers due to erosion of the soil~\cite{varghese2017power}. The work of Shihavuddin et al.~\cite{shihavuddin2019wind} is closer to ours in that they also use deep neural networks to detect damages in power structures, in their case, wind turbines. However, in their work the detection targets are more clearly isolated and delimited, and usually contrast strongly against the white color of the turbines. This makes it a more suitable problem for a bounding-box detector. In our case, this is not possible due to several factors, including the lower visibility of many damages, their clustering and complicated delimitation, their relative lack of conspicuousness within the structures or their similarity in some cases to patches of dirt. These make the detection harder and require more complex techniques to address them.

The contributions of this paper can be summarized as follows:

\begin{itemize}
    \item We propose a solution to the problem of automation of the vision-based inspection of damages in power transmission towers.
    \item We develop techniques to improve the training results despite of the scarcity and ambiguity of the available data. Particularly, we devise a sophisticated augmentation method tailored for the problem at hand and the chosen network, and analyze the best way to deal with the input image size.
    \item We evaluate the viability of our approach as a solution to the problem of damage inspection. Concretely, we propose metrics suitable for our goal and complement them with a qualitative analysis of the results.
\end{itemize}

Additionally, with our work we set the basis for more-advanced damage inspection tasks. Particularly, thanks to the good results achieved in the segmentation of the damages, it is possible in many cases to determine the area of the damaged components, as explained in~\autoref{sec:discussion}. This opens the door to possible future work to further the extent of automation in this process.

\section{\uppercase{Dataset and Methodology}}
\label{sec:dataset_and_methodology}

In this section, we will present the dataset we acquired for our experiments, as well as the methodology used in developing our solution.

\subsection{Dataset}
\label{ssec:dataset}

\begin{figure}[t]
    \begin{center}
        \includegraphics[width=\columnwidth]{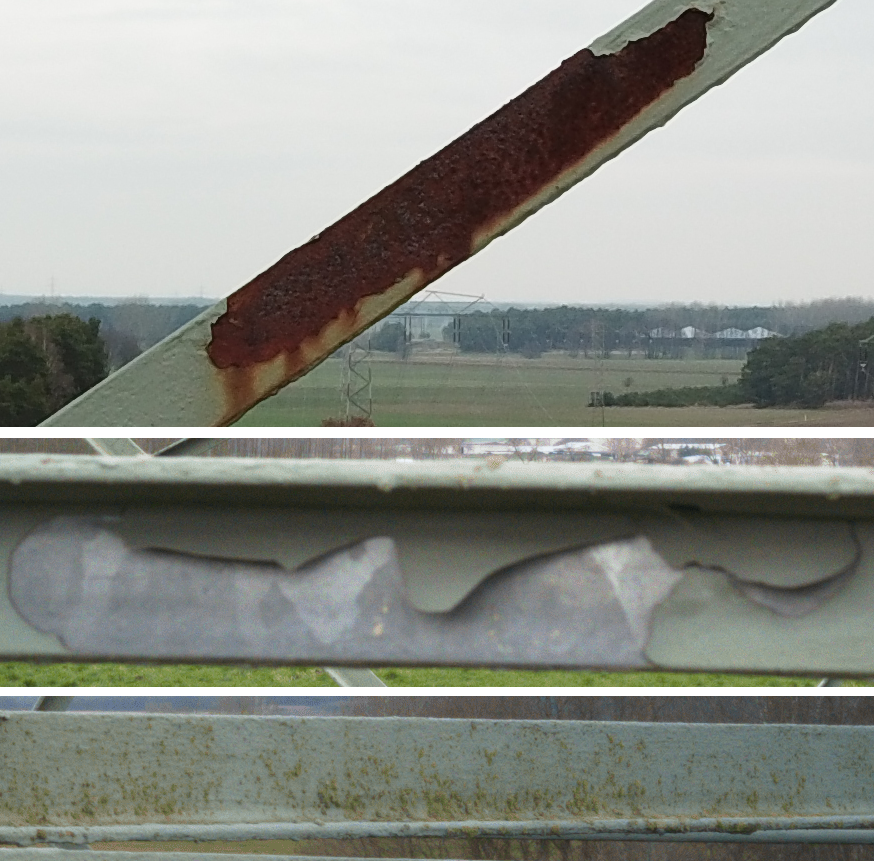}
    \end{center}
    \caption{Examples of a corrosion damage, a missing painting damage and a patch of dirt, from top to bottom.}
    \label{fig:damages_examples}
\end{figure}

The dataset used in our experiments consists of images collected on-site by means of drone-mounted cameras. These drones were remotely controlled by professional pilots that manually flew them around a number of power transmission towers. The images consist mostly of frontal shots of the structures with a line of view roughly parallel to the ground plane, taken from different distances. Since the damages that have been spotted are usually repaired quickly, the towers with actual examples of damages are extremely scarce. This further limits the amount of usable images obtained and creates a class imbalance where most of the image content is background, underscoring the necessity of a method to deal with the small amount of data available. Some images of structures without any issues have also been taken as a reference.

As for the damages, they are quite diverse in appearance, which further complicates their automatic detection. Apart from substantial differences in size, shape and lighting conditions, we have recognized some other intra-class variations. The most common type of damage in our dataset is corrosion, where the layer of paint has completely fallen off and the metal underneath it has become rusty. Another type of damage is the lack of paint, which leaves the metal exposed to the elements and corrosion. It is also possible that only the outermost layer of paint is missing, revealing layers underneath it of a different color but not yet the metal. All of these can be present as either big patches or small dots and on different components, adding yet another degree of intra-class variability. Hence, fewer examples are available for learning for each of these types. Moreover, patches of dirt are also present on these structures, which can easily be mistaken for damages even by an expert labeler. Examples can be seen in~\autoref{fig:damages_examples}.

\begin{figure}[t]
    \begin{center}
        \includegraphics[width=\columnwidth]{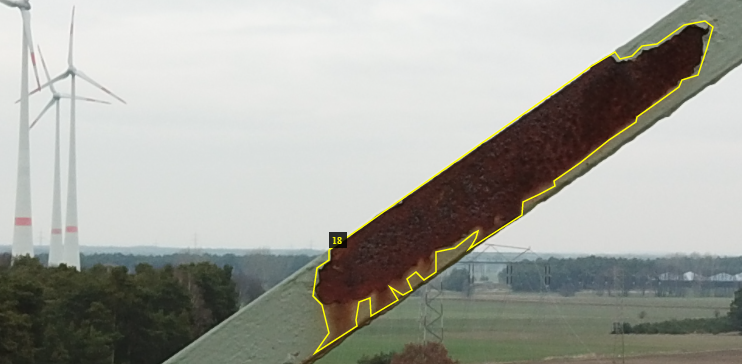}
    \end{center}
    \caption{Example of the polygon-based annotations as labeled by an expert.}
    \label{fig:polygon_annotation}
\end{figure}

The dataset consists of a total of 165 high-resolution pictures, most of them of 4000x2250 pixels, with other similar resolutions also present in smaller numbers. Of these, 20 pictures show structures in a good condition, without any damages. We divided these 165 images randomly into a training set, with 142 images, and a validation set, with 23, approximately 14\% of the total. After this division, we proofed that different damage types had representation in the validation set. The same damages are not present in different pictures. These images have been manually labeled by an expert in the field using the VGG Image Annotator (VIA)~\cite{dutta2019vgg}. The labels consist of polygons that approximate the perimeter of the detection targets, as seen in~\autoref{fig:polygon_annotation}.

\subsection{Instance Segmentation}
\label{ssec:instance_segmentation}

\begin{figure}[t]
    \begin{center}
        \includegraphics[width=\columnwidth]{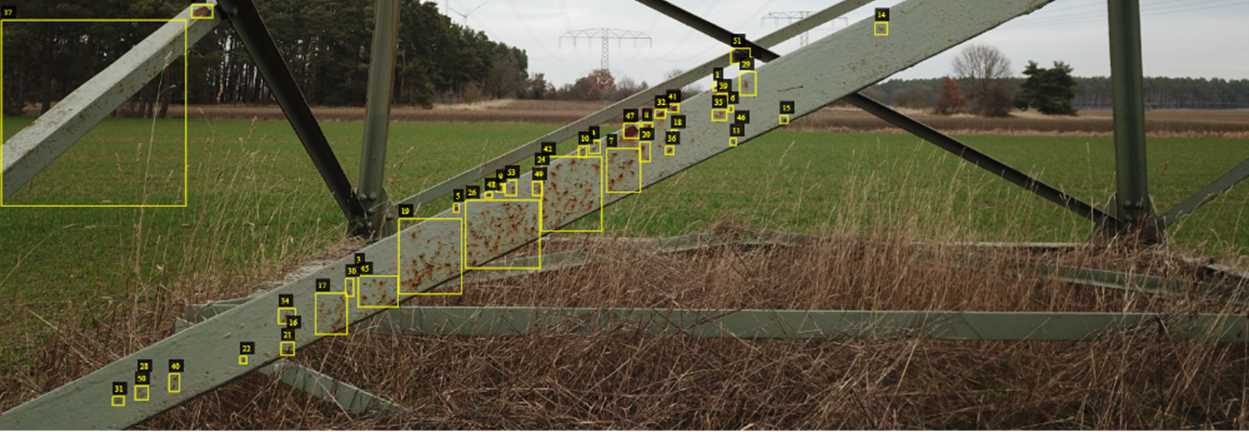}
    \end{center}
    \caption{Examples of difficult cases for labeling with rectangular bounding boxes. Labeling of pixel regions is more suitable in such cases.}
    \label{fig:bad_bounding_boxes}
\end{figure}

In the field of computer vision, there are several tasks related to the detection of objects in images. Some of the most common ones are bounding-box detection, semantic segmentation and instance segmentation. The task of bounding-box detection consists in determining the smallest rectangular box that contains each distinct object of interest in the image. On the other hand, in semantic segmentation one class label is assigned to each pixel, determining what kind of an object it corresponds to, without distinguishing individual objects. Instance segmentation is an intermediate approach in which each individual object of interest in the image is located and the pixel region it occupies is determined.

Although the labels required by bounding-box detection are simpler to produce than those for segmentation, in our dataset many damages are hard to isolate using rectangular boxes, as shown in~\autoref{fig:bad_bounding_boxes}. Especially with such a small amount of examples, it is a difficult task for a neural network to distinguish the relevant content that should be learnt from each box in such cases. This was our experience when trying the Faster-RCNN network~\cite{girshick2015fast}.

With a more time-consuming pixelwise labeling, the damages are more clearly defined. However, in the case of semantic-segmentation, the individual classification of each pixel is a very difficult task to perform due to the strong class imbalance (in average around 99\% of the pixels are background) and, once again, the very limited amount of examples. This was the case for several networks~\cite{ronneberger2015u,zhao2017pyramid,chao2019hardnet,mshahsemseg} we tried, even after modifying their loss functions to compensate for the class imbalance.

An approach based on instance segmentation yielded the best initial results and, thus, is the chosen approach for our experiments. It is introduced in the following section.

\subsection{CenterMask}
\label{ssec:centermask}

In our experiments, we use the CenterMask~\cite{lee2020centermask} architecture for instance segmentation due to its good performance on relevant benchmarks. It consists of three main components:

\begin{itemize}
    \item An interchangeable backbone for feature extraction.
    \item A detection head based on the Fully Convolutional One-Stage Object Detection (FCOS)~\cite{tian2019fcos} architecture.
    \item Spatial Attention-Guided Mask (SAGM) for the segmentation of the detected objects.
\end{itemize}

\begin{figure}[t]
    \begin{center}
        \includegraphics[width=0.8\columnwidth]{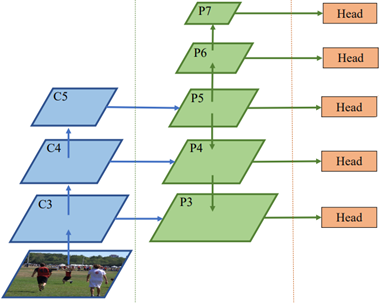}
    \end{center}
    \caption{FCOS feature pyramid (image from original paper). The outputs of applying successive convolutional layers of the backbone to the input are used to generate the different levels of feature maps. From the last one, which constitutes the feature map $P_5$, further convolutions are applied to obtain $P_6$ and $P_7$. In the opposite direction, $P_5$ is scaled successively and combined with the previous levels of outputs to obtain $P_4$ and $P_3$. Predictions are made directly from each of these feature levels, which are responsible for predicting targets of different sizes.}
    \label{fig:feature_maps}
\end{figure}

The architecture of the FCOS detection head is relevant to our data augmentation process due to its feature pyramid system, depicted in~\autoref{fig:feature_maps}, based on the concept proposed in~\cite{lin2017feature}. In the following section, we explain how this design affects our data augmentation process. As for the backbone, we use VoVNetV2~\cite{lee2020centermask} as proposed by the authors.

In our experiments, we use two different classes, damage and dirt, everything else being considered background. The reason we have included dirt as a class is the fact that, in many cases, it can be visually similar to damages, but it is not relevant for the maintenance. In order to avoid raising false alarms about damages, we choose to explicitly learn the difference between both classes.

We have developed different techniques that we applied to the network in order to improve the training. Those yielding the best results in our experiments are discussed in the following subsections.

\subsection{Data Augmentation}
\label{ssec:data_augmentation}

\begin{figure}[t]
    \begin{center}
        \includegraphics[width=0.9\columnwidth]{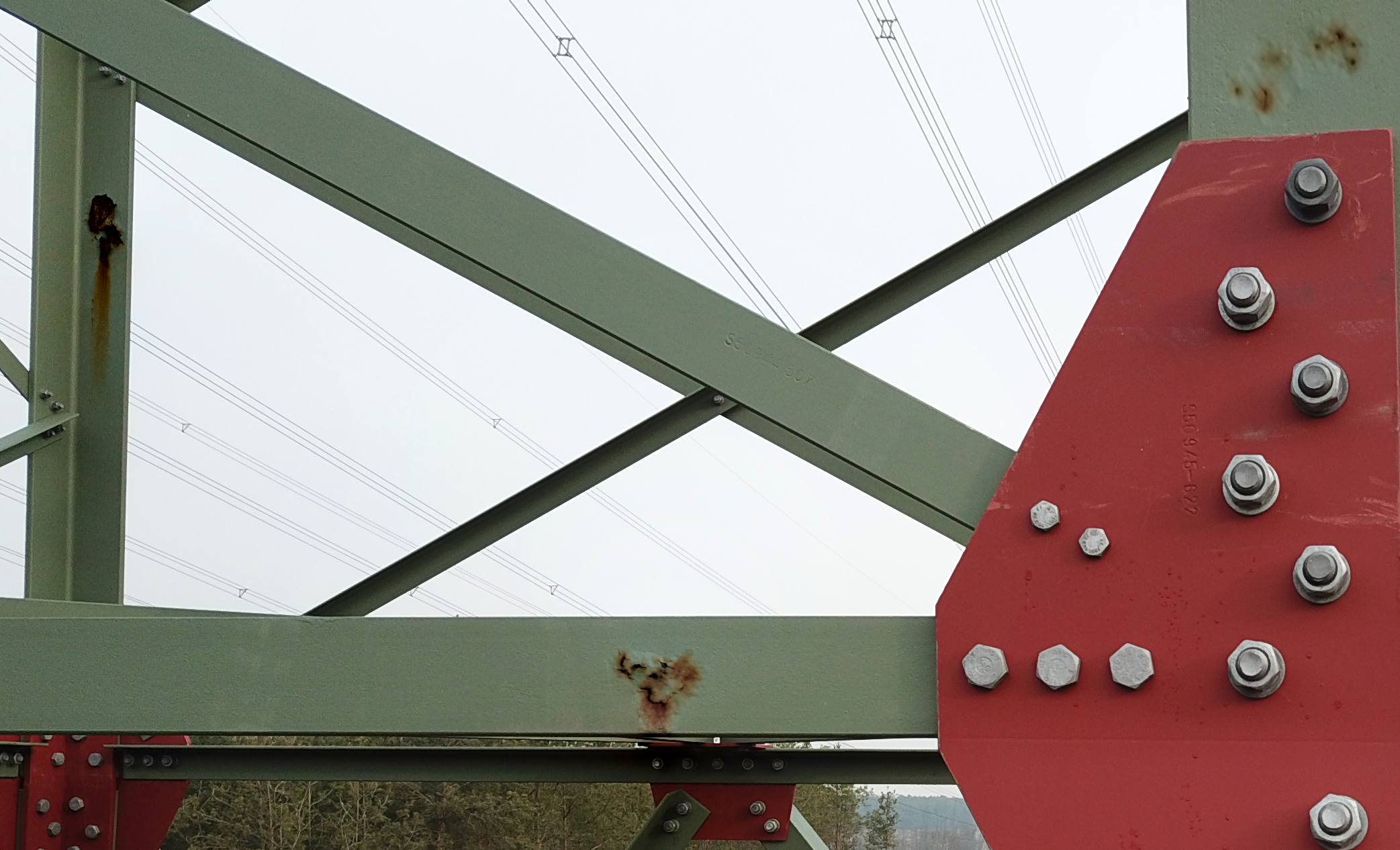}
    \end{center}
    \caption{In this image, three corrosion damages from other pictures have been pasted onto undamaged components using the Poisson Image Editing~\cite{10.1145/882262.882269} technique.}
    \label{fig:poisson_blending}
\end{figure}

One of the focal points of this paper is dealing with the extreme scarcity of training data. One way to mitigate this issue is the use of data augmentation. To this end, different techniques have been tried, ranging from simple framework-supported augmentation techniques to a more sophisticated system we developed for this purpose. As for the basic, framework-supported techniques, we applied to the input images randomized scaling to a given set of short-edge sizes, random horizontal flipping and random cropping. While these simpler methods already increase the diversity of the training data received by the network, the composition of the images remains mostly unchanged. Addressing this issue, we used an augmentation method based on Poisson Image Editing~\cite{10.1145/882262.882269}. In this method, some damages are manually segmented and seamlessly pasted using the aforementioned technique onto undamaged parts of structures in other images, increasing the dataset diversity. An example of this is shown in~\autoref{fig:poisson_blending}. In this method, however, the synthetic images must be manually generated one by one, selecting suitable spots for the chosen damages. We also propose a more sophisticated augmentation method that allows to automatically generate high amounts of synthetic training images, designed after a detailed analysis of the characteristics of the available dataset. In this dataset, we have identified the following issues that need to be addressed:

\begin{itemize}
    \item Severe scarcity of images.
    \item Scarce detection targets for training, especially for each intra-class variation.
    \item Invariability of the appearance and surroundings of these examples.
    \item Fuzzy description of what constitutes a damage yielding unclear examples.
    \item Significant numeric predominance of small, borderline examples over more relevant damages.
\end{itemize}

Additionally, as explained in the \hyperref[ssec:centermask]{previous section}, different feature layers are responsible for predicting objects of different sizes, which means that examples of each type of damage are required at different scales as well.

In order to address these issues more specifically, we have resorted to a method that augments the most relevant examples for the generation of additional synthetic training data. The process of generation of this data is described in the following.

As a first step, a manual selection of the “best examples” from the available training data is performed. The goal here is to focus the training on the most relevant information. Hence, the criteria for determining which ones constitute the best examples are mainly their unambiguity and the coverage of appearance variations. Focusing on these examples should reduce the impact of the numerous small or unclear labels that dominate the images. Increasing the relevance of the good examples this way has yielded much better results than a different approach we explored based on merely reducing the influence of the majority of small, lower-quality examples through a weighting system. The latter was, therefore, discarded.

These selected examples are then manually segmented, including some of the surrounding structure for context, but without background. The regions of the crop around the example that are not part of the structure are removed and left transparent. Examples are taken of both damaged and undamaged components, as well as those with patches of dirt, for a complete coverage.

As a next step, these selected examples are used to generate synthetic images. In order for the network to focus on the examples and ignore the areas of the image that are not relevant, we have decided to use randomized backgrounds. To this end, a collage is made using random images from the Common Objects in Context (COCO) dataset~\cite{lin2014microsoft}, which is then used as a background over which our examples will be pasted. Once the background is generated, a random number of cropped examples are chosen from the different categories. These are processed and pasted onto the background. This processing is aimed at optimizing the quality of the training, and includes the following steps:

\begin{itemize}
    \item \textbf{Random position of the example within the image.} This helps reinforce the translational invariance of the network.
    \item \textbf{Random rotation of the example.} This increases the variation in appearance in our dataset and helps provide rotational invariance.
    \item \textbf{Random cropping of the surroundings of the target.} This makes the learning process less dependent on the exact structure that surrounds the targets, while still allowing us to preserve some of it for context, since we want the network to learn that the damages will always be on the structure.
    \item \textbf{Random scaling.} Each of the examples is individually scaled before being pasted onto the background in order to maximize the size diversity for each type of damage. This is especially relevant in our approach given the architecture of the network, in which different layers are responsible for predicting targets of different sizes. This is expected to make examples of each instance available for learning in different layers.
\end{itemize}

\begin{figure}[t]
    \begin{center}
        \includegraphics[width=\columnwidth]{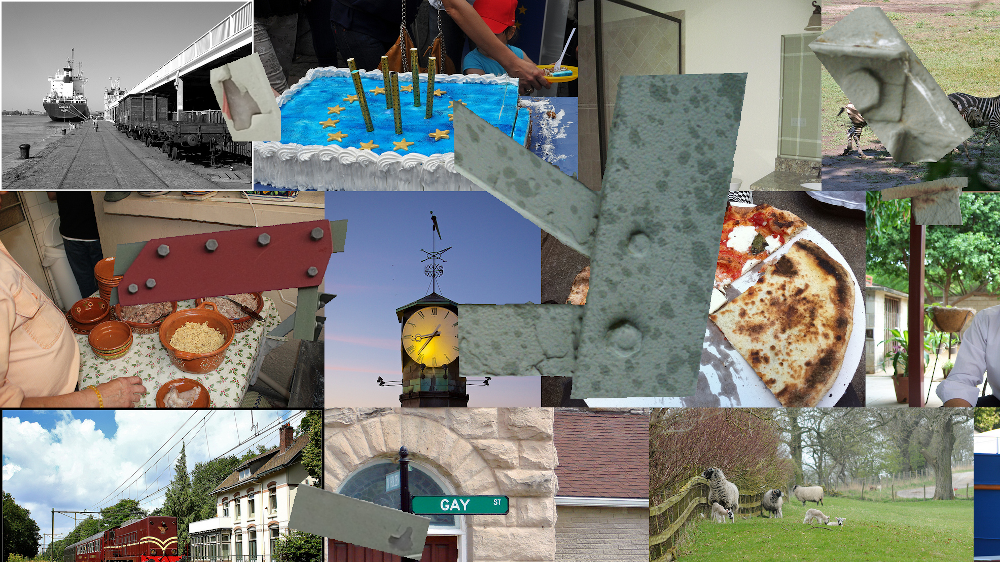}
    \end{center}
    \caption{Example of augmented data. Over the background, composed of random images from COCO dataset, there are six augmented examples: one of them from the class "dirt", three of them from the class "damage", and two of them of undamaged structures that are to be considered background.}
    \label{fig:augmentation_example}
\end{figure}

This means that each synthetic image may have a number of both positive and negative examples with multiple controlled appearance variations, as seen in \autoref{fig:augmentation_example}. The combination of real training data with synthetically generated images has proven a very useful approach to extract the most amount of information from our small dataset, as shown in the results section.

\subsection{Input Image Subdivision}
\label{ssec:input_image_subdivision}

The images available in our dataset are of a very high resolution, which makes it unfeasible for the network to process them directly. Hence, an important decision is how to deal with the size of the input.

One approach would be to perform a downsampling of the input images until they reach a manageable size for the network. Experimentally, we have found that scaling both dimensions of the images to approximately 50\% of their original values makes it feasible for the network to process them on common consumer graphics cards with 11 GB RAM. Doing this, however, implies a relatively high loss of information, although most of the relevant damages are still clearly recognizable to the human eye.

Another option consists in splitting the image into smaller parts that can then be processed in nearly original size as separate inputs. Expected downsides of this approach include the extra processing time required and potential loss of context by cropping. However, the visual information would be preserved and, in such a small dataset, feeding image subsections separately in random order could come as an advantage to artificially increase the training diversity. This option has been implemented as a “sliding window” approach, dividing the input images vertically and horizontally in overlapping regions of the same size. We expect this overlap to help mitigate the loss of context due to cropping, since the same region is present in different windows with different surroundings, preventing potentially misleading crops of targets on the edges. As a final step, the predictions from different windows are offset and the overlapping ones aggregated to be merged into a single image. Experimentally, we have found this approach to offer better results than downsampling of the input.

\section{\uppercase{Results}}
\label{sec:results}

In this section, we propose different performance metrics and discuss their suitability to our problem. These metrics are used for a quantitative analysis of the results. This is complemented by an additional qualitative analysis, taking into account the particular characteristics of our data. We take the network trained without using any of the proposed techniques as a baseline to compare our methods.

The following two performance metrics have been used to analyze the quality of the results:

\begin{itemize}
    \item Precision vs. recall. In this case, we define precision as the ratio of predictions that have any overlapping with a detection. This accounts for our goal to detect all damages, not to accurately segment them. As for recall, we define it as the ratio of the labeled damages that have been detected, again taking any overlapping with a prediction as a detection. This metric gives a rough idea of how many of the total damages have been detected and how many false detections have been made. Its main flaw is the dominance of small damages, which might not be so relevant.
    \item Intersection over union (IoU). This metric is calculated as the ratio between the area of the intersection of predictions and targets and this of their union. The downside of this metric is that it tends to assign low scores when the surface has not been predicted with a very high degree of accuracy, which in our case is an ill-posed problem given that, as previously mentioned, the boundaries of damages are many times unclear even to an expert annotator. Moreover, accurate segmentation is not the goal of this system. On the other hand, it has the advantage of determining how much of the labeled surface has been predicted correctly, reducing the influence of less-relevant, tiny damages. This metric has been computed over the total labeled/predicted area across the validation set instead of per image, which is expected to give a more accurate representation of the success rate in such a small dataset.
\end{itemize}

In the following, we will analyze the results using both the proposed metrics and a qualitative discussion in order to provide a more complete insight.

\begin{figure}[t]
    \begin{center}
        \includegraphics[width=\columnwidth]{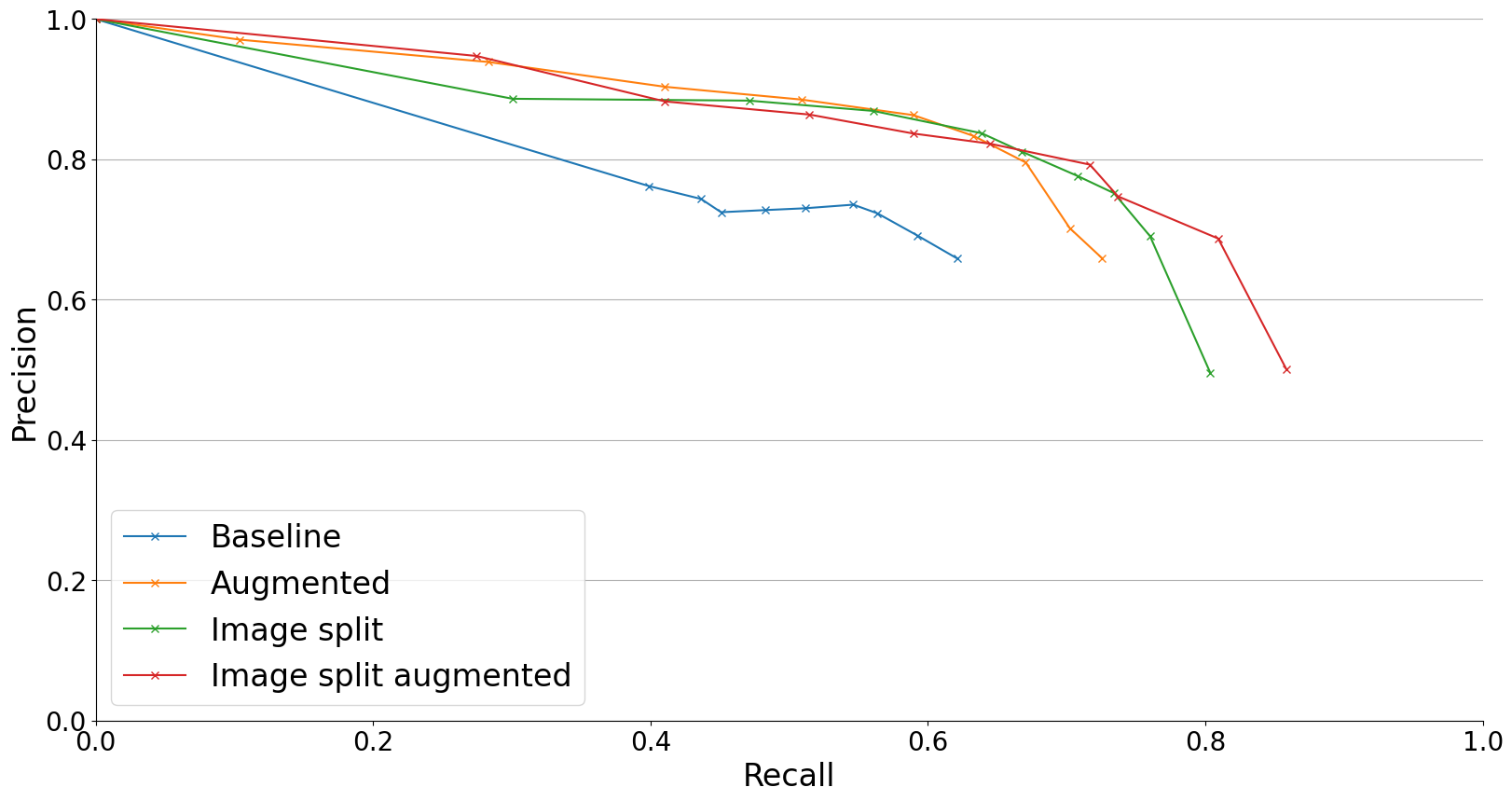}
    \end{center}
    \caption{Precision-recall curves for the different approaches. Baseline (blue) corresponds to the network trained without using any of the proposed improvement techniques, Augmented (yellow), to the use of our sophisticated augmentation technique, Image Split (green), to the use of the proposed subdivision of input images, and Image Split Augmented (red) to the combination of these two techniques.}
    \label{fig:results_pr_curve}
\end{figure}

\begin{figure}[t]
    \begin{center}
        \includegraphics[width=\columnwidth]{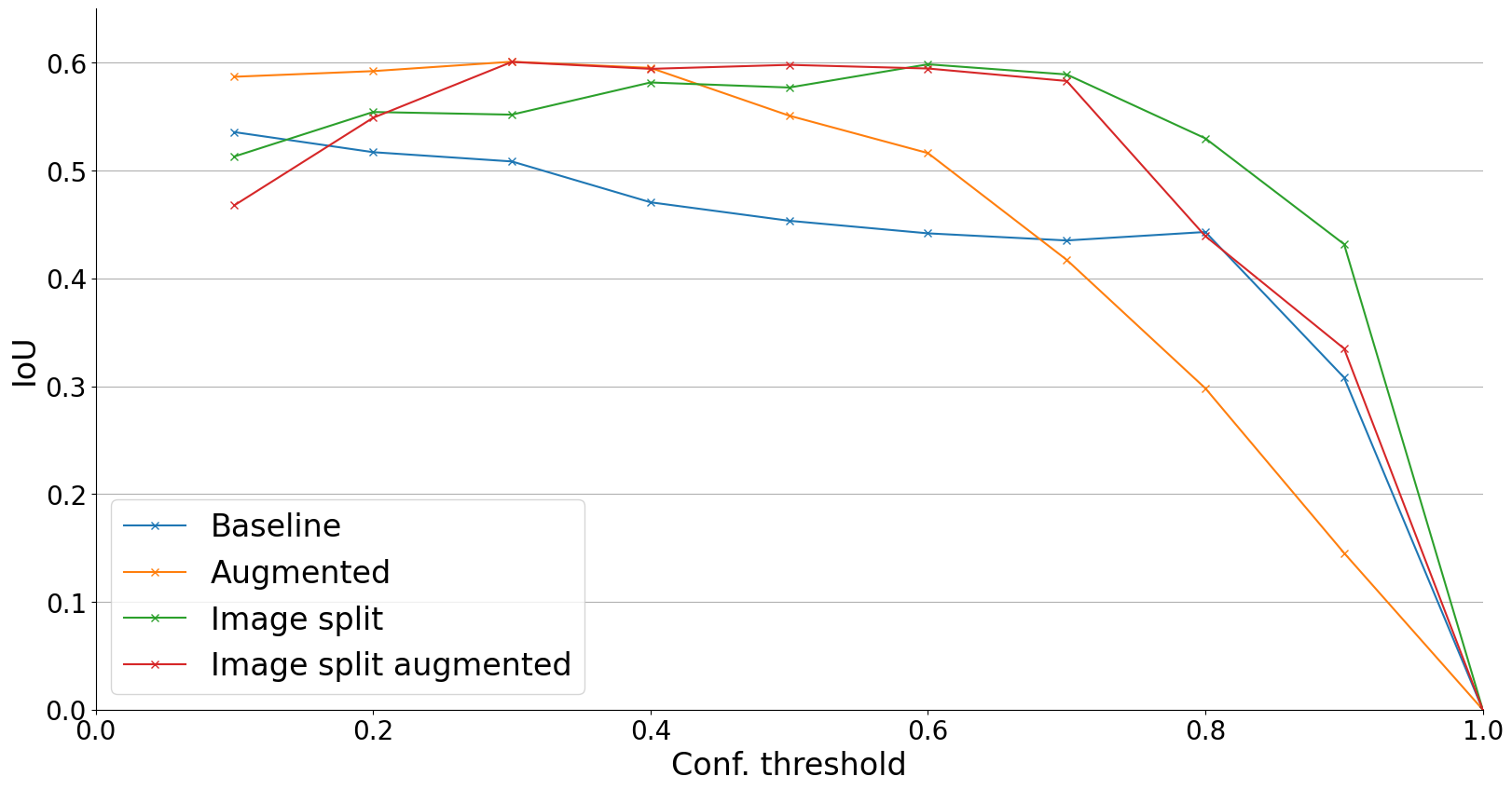}
    \end{center}
    \caption{IoU values for each confidence threshold used for the different approaches. Baseline (blue) corresponds to the network trained without using any of the proposed improvement techniques, Augmented (yellow), to the use of our sophisticated augmentation technique, Image Split (green), to the use of the proposed subdivision of input images, and Image Split Augmented (red) to the combination of these two techniques.}
    \label{fig:results_iou}
\end{figure}

\newlength\q
\setlength\q{\dimexpr .25\columnwidth -1.8\tabcolsep}

\begin{table}
    \centering
        \caption{Maximum IoU values for each method in \autoref{fig:results_iou}.}
        \begin{tabular}{ |p{0.8\q}|p{\q}|p{0.8\q}|p{1.2\q}| }
            \hline
            Baseline & Augmented & Image Split & Image Split Augmented \\\hline
            53.56\%  & 60.09\%   & 59.85\%     & 60.07\% \\\hline
        \end{tabular}
        \label{tab:results_iou_top}
\end{table}

\begin{figure}[t]
    \begin{center}
        \includegraphics[width=\columnwidth]{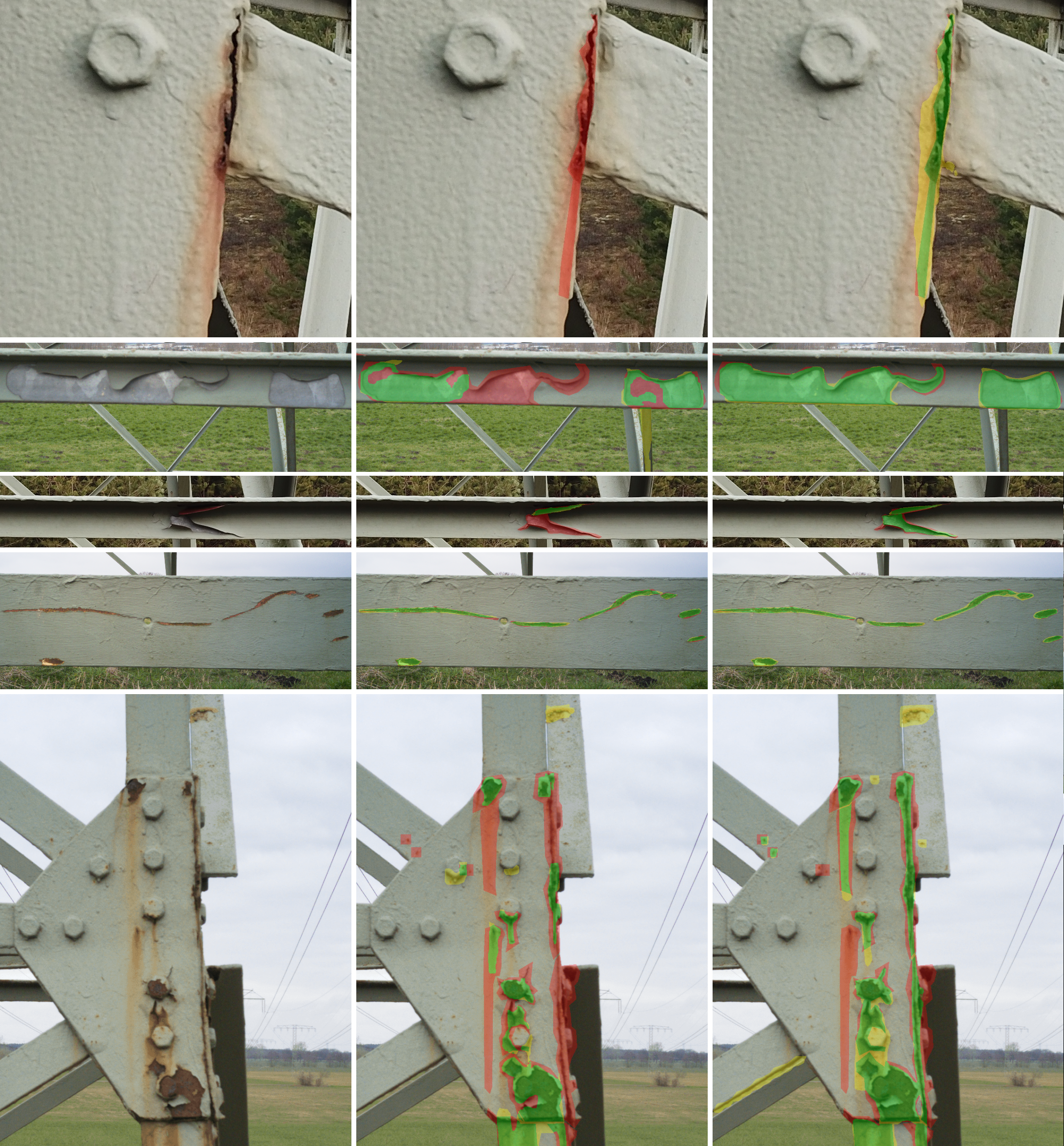}
    \end{center}
    \caption{Prediction examples. The left column corresponds to parts of original images showing damages, the middle one to the baseline approach, and the right one to the approach using augmentation and image splitting. The correctly detected areas are colored green, the undetected areas labeled as damages, in red, and the false positives, in yellow.}
    \label{fig:results_view}
\end{figure}

\autoref{fig:results_pr_curve}~shows the first of the proposed metrics, precision vs. recall, applied to the different approaches. \autoref{fig:results_iou}~shows the second one, with its maximum values displayed in \autoref{tab:results_iou_top}. For both metrics, it is clear that the baseline is overcome by all 3 of our approaches, particularly so for the approaches including our augmentation method. The latter seem to be better at suppressing false positives, with their best results being for lower confidence thresholds. This might be due to better examples being emphasized in our training method, ignoring the bad ones. On the other hand, the approaches using input image splitting have a higher maximum recall, which was expected due to the lower loss of resolution in the input. The method including both techniques seems to offer a good compromise solution. \autoref{fig:results_view} shows some detection examples for the baseline method and our proposed approach (the combined method).

From these examples, we can draw several conclusions. First, in some cases detections marked as false positives are unclear and could very well have been labeled as damages by a different annotator. The exact borders of the damages, which often affect substantially the IoU score, could also in many instances be subject of discussion. Other than this, we see that most of the clearly visible damages have been segmented with a reasonable level of accuracy by all methods. However, some of the intra-class variations have proven to be harder to detect, particularly those cases in which the coating has fallen off, showing the metal below, which is of a similar color. For these, it is possible to see that our proposed best-solution shows clearly better results than the baseline approach, generalizing better for damage types for which there are less examples or that are visually more challenging to detect.

\section{\uppercase{Discussion}}
\label{sec:discussion}

\begin{figure}[t]
    \begin{center}
        \includegraphics[width=0.9\columnwidth]{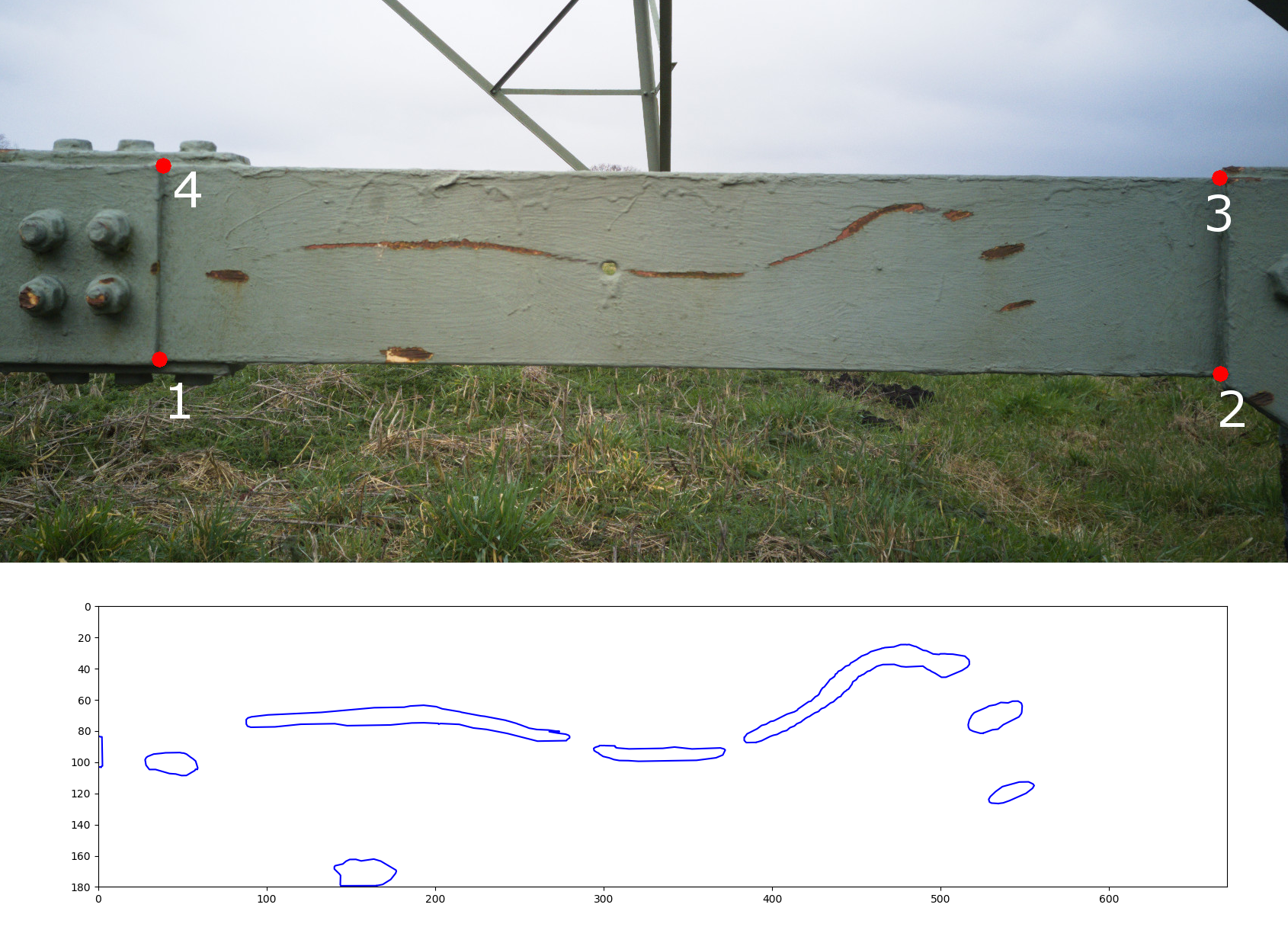}
    \end{center}
    \caption{On the top, a reference component with a damage is shown. Below, the real measurements of the detected damages.}
    \label{fig:damage_measure}
\end{figure}

As discussed in the results section, most of the clearly visible damages in the validation set have been recognized by the system. After having applied the proposed techniques to improve the training, this includes most of the challenging intra-class variations, such as the lack of paint without corrosion.

As for the undetected damages, for many of them there are different reasonable labeling possibilities and we expect that different annotators would generate different labels for such cases, perhaps with a relatively low IoU score between them. Many others are due to poor visibility. We expect that a denser image coverage of the structures would yield better points of view that would help solve this.

Given these results, we believe that the system would be suitable for its purpose, namely minimizing human interaction in the inspection of structures through automation. Although ideally its performance could be improved by increasing the amount of training data, the relatively high detection rate and reduced number of false positives achieved are expected to guarantee a relevant speedup with respect to human-only inspection. As a matter of fact, the good segmentation results make it possible to determine the area of damaged components in many cases. In an experiment, given the dimensions of a reference component we calculated the damaged area, as shown in~\autoref{fig:damage_measure}.

\section{\uppercase{Conclusions}}
\label{sec:conclusions}

In this paper, we have proposed a system for automated detection of damages in power transmission towers using drone images. We have successfully overcome the limitations  of the scarcity of training data and the inherent ambiguity of part of it. These challenges have been tackled using an instance segmentation network and diverse techniques to optimize the training suited to the particular conditions of the available data. Furthermore, we have evaluated the results obtained by the proposed methods, both quantitatively and qualitatively, and compared them to the baseline. In doing so, we reached the conclusion that, not only do these techniques improve the results, but the resulting system shows a promising performance that would make it suitable for our goal of automation. Thanks to our sophisticated augmentation, damages of different types are detected accurately even if they are underrepresented or subjectively labeled in the training data. We expect that the usage of this system will help reduce the human input needed and substantially speed up the whole process.

\bibliographystyle{apalike}
{\small
\bibliography{example}}

\end{document}